\documentclass[10pt,twocolumn,letterpaper]{article}

\usepackage{cvpr}
\usepackage{times}
\usepackage{epsfig}
\usepackage{graphicx}
\usepackage{amsmath}
\usepackage{amssymb}
\usepackage{math}

\cvprfinalcopy 

\def\cvprPaperID{1552} 

\ifcvprfinal\fi
\begin{document}

\title{Articulated Shape Matching Using Laplacian Eigenfunctions and\\ Unsupervised Point Registration}

\author{
\begin{tabular}{ccccc}
Diana Mateus\thanks{is a {\sc p}h.{\sc d} fellow of Marie-Curie Early Stage Training action Visitor.}
& Radu Horaud
& David Knossow
& Fabio Cuzzolin\thanks{{\sc f.c.} is a post-doctoral fellow of Marie-Curie research training network Visiontrain.}
& Edmond Boyer
\end{tabular}
\\
INRIA Rh\^one-Alpes\\
655 avenue de l'Europe, Montbonnot Saint-Martin\\
38 334 Saint-Ismier Cedex France\\
}
\maketitle


\begin{abstract}
\vspace{-3mm}
Matching articulated shapes represented by voxel-sets reduces to maximal sub-graph isomorphism when each set is described by a weighted graph. Spectral graph theory can  be used to map these graphs onto lower dimensional spaces and match shapes by aligning their embeddings in virtue of their invariance to change of pose. Classical graph isomorphism schemes relying on the ordering of the eigenvalues to align the eigenspaces fail when handling large data-sets or noisy data. We derive a new formulation that finds the best alignment between two congruent $K$-dimensional sets of points
by selecting  the best subset of eigenfunctions of the Laplacian matrix. 
The selection is done by matching eigenfunction \emph{signatures} built with histograms, and the retained set provides a smart initialization for the alignment problem with a considerable impact on the overall performance. Dense shape matching casted into graph matching reduces then, to point registration of embeddings under orthogonal transformations; the registration is solved using the framework of unsupervised clustering and the EM algorithm. Maximal subset matching of non identical shapes is handled by defining an appropriate outlier class. Experimental results on challenging examples show how the algorithm naturally treats changes of topology, shape variations and different sampling densities.
\end{abstract}
\vspace{-5mm}
\begin{figure}[ht]
\centering
\includegraphics[clip=true, viewport= 20  100 1000 900, keepaspectratio=true,width=0.8\columnwidth]{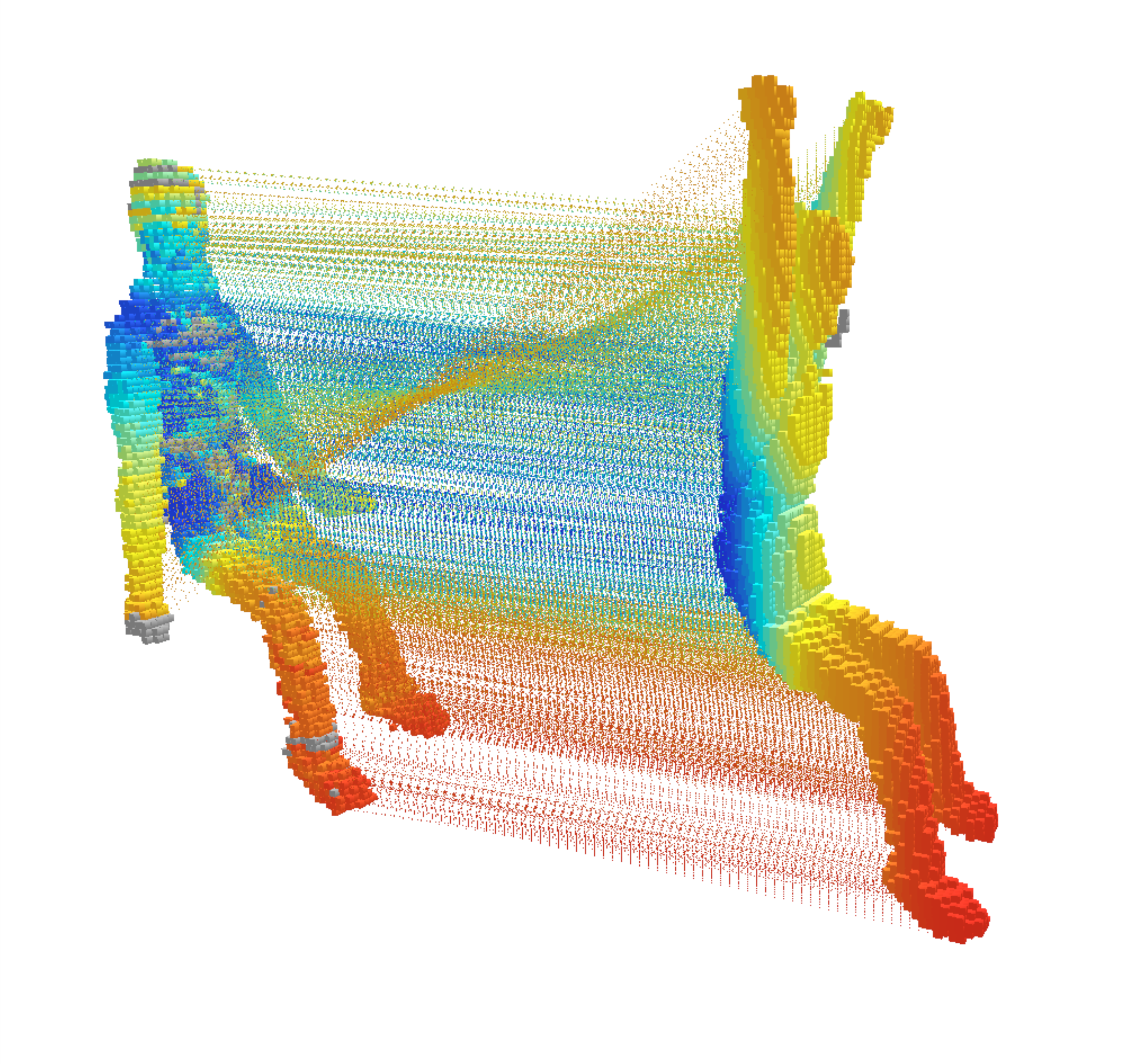}
\caption{Two articulated shapes, of 12577 and 12346 voxels, matched with our method. The two voxel-sets are embedded in a 7-dimensional space. Using eigenfunction alignment to initialize the 7$\times$7 orthogonal trasformation, EM converges after 9 iterations. There are only 90 unmatched voxels(in grey).}
\vspace{-5mm}
\label{fig:matching-example}
\end{figure}


\section{Introduction}
\vspace{-3mm}
Shape matching is a central topic in many areas of computational vision, such as object modeling, motion tracking, object and action recognition. One of the main goals of shape matching is to find dense correspondences between the representations of two objects (as illustrated on Figure~\ref{fig:matching-example}.
In the recent past there has been tremendous interest in both \hbox{2-D} and \hbox{3-D} shape matching. In spite of these efforts, {\em the problem of matching \hbox{3-D} articulated shapes} remains very difficult, mainly because it is not yet clear how to choose and characterize the group of transformations under which such shapes should be studied \cite{EladKimmel2003}, \cite{ChuJenkinMataric2003}, \cite{anguelov-davis:nips2004}, \cite{JainZhang2006}.
One possible approach, among many others, is to represent shapes by locally connected sets of points, i.e. \emph{sparse} graphs, and to use spectral embedding methods in order to isometrically map these graphs onto a lower dimensional space. As a result, a dense match between shapes can be found through the alignment of their embeddings. 
Umeyama \cite{Umeyama88} proposed a solution to the weighted graph matching problem based on eigendecomposition of the graphs' adjacency matrices, and Scott and Longuet-Higgins \cite{ScottLonguetHiggins91} who used the Gaussian proximity matrix to match sets of rigid points, pioneered this research area. The proximity matrix was also used in \cite{ShapiroBrady92} while \cite{CarcassoniHancock2003a,CarcassoniHancock2003b} extended the Gaussian kernel to other robust weight functions and proposed a probabilistic framework for point-to-point matching.
The main drawback of these embedding techniques is that they cannot be easily extended to articulated shapes. Also they are quite sensitive to noise, to outliers, as well as to discrepancies between the two point-sets (e.g. missing parts, deformations, etc.). 

To extend the the use of spectral graph methods to the unsupervised representation and matching of articulated shapes, one can use local isometries as invariants preserved by articulated motion (except in the vicinities of the joint). Methods such as Laplacian Eigenmaps \cite{BelkinNiyogi2003}, generally used for dimensionality reduction, are explicitly conceived to find a mapping that best preserves such pairwise relationships. 
Indeed, the Laplacian matrix \cite{Chung97} encodes {\em local} geometric and topological properties of a graph, and hence it is well suited for describing articulated objects. Belkin and Niyogi \cite{BelkinNiyogi2003} provide a theoretical justification for combining the Laplacian operator with the Gaussian (or heat) kernel. Laplacian embedding has strong links with spectral clustering  \cite{NgJordanWeiss2002} and with other local-based embedding methods such as LLE \cite{RoweisSaul2000}. In \cite{JainZhang2006}, the Laplacian embedding is used to embed \hbox{3-D} meshes using a global geodesic distance and to match the embeddings using the ICP algorithm \cite{Zhang94}.

In this paper we propose a new method for matching articulated shapes 
which combines Laplacian embedding with probabilistic point matching, \cite{DewaeleDevernayHoraudForbes2006,LuoHancock2003,WellsIII97}.  We thoroughly justify the choice of the Laplacian for capturing the properties of locally-rigid shapes, for mapping them onto a low-dimensional subspace of the eigenspace, and for performing probabilistic point registration. Problems with the ordering of eigenvalues have been identified before \cite{JainZhang2006} with point-sets of  cardinality $10^2$ . We offer an interesting alternative to eigenvalue ordering, namely the use of eigenfunction histograms. Our new alignment method is particularly relevant when the cardinality of the point-sets is very large (of the order of $10^4$) while the dimension of the embedding space is of the order of $10$. Since we conclude that articulated shape matching is equivalent to point-to-point assignment under an orthogonal transformation, we concentrate on the latter and provide an algorithm that handles point registration in the framework of unsupervised clustering. We claim that, by adding a uniform component to a mixture of Gaussians, it is possible to deal with discrepancies between the shapes and with outliers.

The remainder of this paper is organized as follows. Section \ref{section:problem-statement} establishes a link between graph isomorphism and point registration.
Section \ref{section:laplacian-eigenfunctions} describes a method for aligning eigenfunctions based on their histograms.
Section \ref{section:point-registration} casts the point registration problem in the framework of unsupervised clustering.
Section \ref{section:experiments} describes experiments with several data sets, and
Section \ref{section:conclusions} draws some conclusions and proposes directions for future work.
\vspace{-0.2cm}
\section{Problem statement}
\label{section:problem-statement}
\vspace{-0.2cm}
\subsection{Graph Laplacian operator}
\label{section:laplacian-operator}
\vspace{-3mm}

Given a voxel-set representation of an articulated shape, let $\mathcal{X}=\{X_1,\ldots,X_N\}$ be the set of voxel centers ($X_{i} \in \mathbb{R}^{3}$). It is possible to define a graph  $\mathcal{G}_x$, where each node is associated with a $X_{i}$ and where the local geometry of the set  is used to define the edges. The voxel connectivity together with a notion of local distance between pairs of nodes determines a weighted adjacency matrix $\mm{W}$ with entries:
\begin{equation}
W_{ij}  = \left\{
\begin{array}{l}
\exp (-d^2(i,j)/\nu^2) \text{ if $i\in \mathcal{N}(j)$}\\
0 \text{ if $i\notin \mathcal{N}(j)$}\\
0 \text{ if } i=j
\end{array}
\right.\\
\end{equation}
where $d(i,j)$ denotes the \emph{local} shortest-path distance between points associated with nodes $i$ and $j$, $\nu$ is a scalar, and $\mathcal{N}(j)$ is the set of neighbors of $j$. 
In order to analyze some of the geometric and topological properties of these graphs (e.g., patterns of connectivity), we can use spectral graph theory \cite{Chung97}, and in particular, the normalized graph Laplacian $\mm{L}=\mm{D}^{-1/2}\mm{W}\mm{D}^{-1/2}$. Here $\mm{D}$ stands for the diagonal degree matrix whose elements verify $D_{ii} = \sum_j W_{ij}$. 



\subsection{Spectral graph matching}\label{section:spectralgraphmatch}
\vspace{-3mm}

Consider another articulated shape $\mathcal{Y}=\{Y_1,\ldots,Y_M\}$ and let for the time being {\small$N=M$}. From $\mathcal{X}$ and $\mathcal{Y}$ we build two graphs $\mathcal{G}_x$ and $\mathcal{G}_y$ and two adjacency matrices $\mm{L}_x$ and $\mm{L}_y$. Matching these graphs is the problem of finding a {\small$N \times N$} permutation matrix $\mm{P}$ minimizing the function:
\begin{equation} \label{eq:j}
J(\mm{P})=\|\mm{L}_x - \mm{P} \mm{L}_y \mm{P}{\tp}\|^2.
\end{equation}
According to Umeyama's theorem \cite{Umeyama88},  the solution can be found through the alignment of the graphs' eigenspaces, as follows. 
Let $\mm{L}_x=\mm{U}_x\mm{\Lambda}_x\mm{U}_x\tp$ and $\mm{L}_y=\mm{U}_y\mm{\Lambda}_y\mm{U}_y\tp$ be the eigendecompositions of the corresponding adjacency matrices, where $\mm{U}_x \text{ and } \mm{U}_y$ are two orthogonal matrices and $\mm{\Lambda}_x=\diag[\lambda_1,\ldots,\lambda_N]$, $\mm{\Lambda}_y=\diag[\delta_1,\ldots,\delta_M]$.
First, the domain of the objective function (\ref{eq:j}) is extended to the space of all orthogonal matrices $\mm{Q}$. This extension is natural because permutation matrices belong to the group of orthogonal matrices. Umeyama's theorem states that {\em if the eigenvalues of $\mm{L}_x$ and $\mm{L}_y$ are distinct and if they can be ordered}, then the minimum of $J(\mm{Q})$ is reached for \hbox{$\mm{Q}^{\ast}=\mm{U}_x\mm{S}\mm{U}_y\tp$}.
The diagonal matrix \hbox{$\mm{S}=\diag[s_1,\ldots,s_N], s\in\{+1;-1\}$} accounts for the sign ambiguity in the eigendecomposition. The matrix $\mm{Q}^{\ast}$ is an alignment of the complete eigenbasis. For two perfectly isometric shapes, $\mm{Q}^\ast$ is also identical to the searched permutation matrix $\mm{P}$ and determines a node-to-node assignment function (acting on the indexes of one node sets) $\omega: \{1,...,M \}\mapsto \{\omega(1),...,\omega(M)\}$. 


One of our main \textit{contributions}, is the extension of this theorem 
in order to solve the maximum sub-graph matching problem through the registration of the graphs' embeddings  under an orthogonal transformation.


\subsection{Graph matching in a reduced eigenspace}\label{section:reduced}

For \emph{large} and \emph{sparse} graphs, Umeyama's theorem holds only {\em weakly}. 
Indeed, one cannot guarantee that the eigenvalues of the Laplacian matrix are all distinct.  Furthermore, the presence of  symmetries in the shape causes some of eigenvalues to have an algebraic multiplicity greater than one. Under these circumstances and due to numerical approximations, it might \emph{not} be possible to properly order the eigenvalues. In the absence of an absolute eigenvalue ordering, Umeyama's theorem is impractical (one would need to find the best out of all possible $N!$ orderings).

One elegant way to overcome this problem, is to reduce the dimension of the eigenspace, along the line of spectral nonlinear reduction techniques. Following the Laplacian eigenmaps scheme \cite{BelkinNiyogi2003},  one can perform a generalized eigendecomposition of the Laplacian matrix $\mm{L}$ and use the  $K$ smallest  eigenvalues\footnote{leaving out the eigenvalue 0 and its unitary eigenvector} with their associated eigenspace ({\small $K\ll N$}). We denote by $\mm{U}_x^{K}$ the {\small $N\times K$} block matrix of $\mm{U}_x$. The columns of $\mm{U}_x^{K}$ correspond to the $K$ eigenvectors associated with the selected eigenvalues, while its rows represent the coordinates of the set $\mathcal{\bar{X}}$ in the embedded $K$-dimensional space:  $\mathcal{\bar{X}}=\{x_1,\ldots,x_N\}$. Since $X_{i}\in \mathbb{R}^{3}$ and $x_{i}\in \mathbb{R}^{K}$, this corresponds to a mapping  $\mathbb{R}^{3}\mapsto \mathbb{R}^{K}$. Respectively, for the second shape $\mathcal{Y}$, we can use the matrix $\mm{U}_y^{K}$ to derive the embedding $\mathcal{\bar{Y}}=\{y_1,\ldots,y_M\}$.

Because it preserves local geometry,  the Laplacian embedding projects a pair of locally isomorphic shapes (true for articulated shapes) onto two congruent point-sets in $\mathbb{R}^{K}$. This condition makes it suitable for Umeyama's framework.
If the $K$ eigenvalues of the reduced embeddings were distinct and reliably ordered, one could directly use Umeyama's theorem and the minimizer of $J(\mm{Q})$ in the reduced eigenspace:
\begin{equation}
\mm{Q}^{\ast}		=	 \mm{U}_{y}^{K}		\mm{S}^{K}		{\mm{U}_{x}^{K}}{\tp} .
\end{equation}
\vspace{-0.7cm} 

Notice that $\mm{S}^{K}$ is now {\small{$K\times K$}}.
If we can not reliably oder the eigenvalues: $\{\lambda_{1},\ldots,\lambda_{K}\}$, $\{\delta_{1},\ldots,\delta_{K}\}$, we need to add a permutation $\mm{P}^{K}$:
\begin{equation}
\mm{Q}^{\ast}=\mm{U}_{y}^{K}\mm{S}^{K}\mm{P}^K{\mm{U}_{x}^{K}}{\tp}.
\end{equation}
Again, $\mm{P}^{K}$ is only {\small $K\times K$}. Let $\mm{R}_{K}=\mm{S}^{K} \mm{P}^K$ and extend the domain of $\mm{R}_{K}$ to all possible orthogonal matrices of size $K\times K$. This is done both, to find a close-form solution and to deal with algebraic multiplicities. As a result:
\begin{equation}
\mm{Q^{\ast}}=\mm{U}_{y}^{K}\mm{R}^{K}{\mm{U}_{x}^{K}}{\tp},
\end{equation}
$\mm{R}^{K}$ works as an orthogonal transformation aligning the \hbox{$K$-dimensional} coordinates of the two point-sets:
\begin{equation}
{\tmm{U}_x^K}{\tp} = \mm{R}^K {\mm{U}_y^{K}}{\tp}.
\label{eq:point-alignment}
\end{equation}
where the column vectors of ${\tmm{U}_x^{K}}{\tp}=[x_{\omega(1)} \ldots x_{\omega(M)}]$ are assigned to the column vectors of ${\mm{U}_y^K}{\tp}= [y_1 \ldots y_M]$, with $\omega: \{1,...,M \}\mapsto \{\omega(1),...,\omega(M)\}$. Under an exact local-isomorphism assumption, the alignment matrix $\mm{R}^{K}$ makes the estimation of $\omega$ trivial (it can be performed by a simple nearest-neighbor method). 
To conclude, we can state the following proposition:

\textbf{Proposition:} Let two articulated shapes be described by two graphs as defined in section \ref{section:laplacian-operator}. Consider the Laplacian embeddings of these two graphs onto a $K$-dimensional space. Solving for the local graph isomorphism problem is equivalent to finding a solution for the point registration problem under the group of orthogonal transformations. Namely, estimate a $K\times K$ orthogonal matrix $\mm{R}^K$ that aligns the reduced eigenspaces and find the trivial assignment  $\mm{\omega}$.


Nevertheless, in real scenarios the two shapes are never locally identical and the number of voxels in each set can be different. To be able to recover the assignment $\omega$, the problem needs to be reformulated as a \emph{maximum sub-graph isomorphism}, i.e.  finding the largest match between sub-graphs in the two shapes.  In practice, this amounts  to relaxing the constraints of  the assignment $\omega$ being strictly one-to-one and  the two graphs having the same number of nodes.  One possible solution is to devise an algorithm that alternates between estimating the $K\times K$ orthogonal transformation $\mm{R}^{K}$, which aligns the $K$-dimensional coordinates of the two points sets, and finding an assignment $\omega$. One may observe that $\mm{R}^K$ belongs to the orthogonal group $O(K)$, therefore this framework is different from the classical rigid point registration, where one seeks for a \hbox{3-D} rotation, i.e. a member of the special orthogonal group $SO(3)$. 

Since an alternation approach may lead to local-minima, we propose a solution to the \emph{maximum sub-shape isomorphism} problem in two stages. First, in section \ref{section:laplacian-eigenfunctions}, we describe a method that estimates the best matrices $\mm{P}^K$ and $\mm{S}^K$, relying neither on eigenvalue ordering nor on exhaustive search, but instead in a \emph{comparison of the eigenfunction histograms}  of the two embeddings. Then, in section \ref{section:point-registration}, we detail a robust point registration method, based on maximum likelihood with latent variables. The algorithm follows the clustering formulation of EM, and is  initialized with the $\mm{P}^K$ and $\mm{S}^K$ obtained in the previous stage.

\section{Alignment using Laplacian eigenfunctions}
\label{section:laplacian-eigenfunctions}
\vspace{-3mm}
In the context of embedded representations of graphs, the columns of matrix $\mm{U}_x^K$  may be viewed as a set of eigenfunctions, each such eigenfunction maps the graph $\mathcal{G}_x$ onto $\mathbb{R}$. Similarly a column of $\mm{U}_y^K$
maps $\mathcal{G}_y$ onto $\mathbb{R}$. For two identical graphs, these eigenfunctions should be identical, up to the ordering of the eigenvalues (which implies an ordering of the column vectors of both $\mm{U}_x^K$ and $\mm{U}_y^K$) and up to a node-to-node assignment. However, empirical evidence points out that eigenvalue ordering is not reliable (hence the presence of matrix $\mm{P}^K$) and we do not have a node-to-node assignment. Let $\vv{u}_{x}^k$ (respectively $\vv{u}_{y}^k$) be the $k^\text{th}$ eigenfunction of the Laplacian associated with the graph $\mathcal{G}_x$ (respectively $\mathcal{G}_y$). $\vv{u}_{x}^k$ can be seen as a column vector, whose histogram $h(\vv{u}_{x}^k)$ is obviously invariant with respect to the order of the vector's components, and thus to the order in which the graph nodes are considered. The histogram can be regarded then as an {\em eigenfunction signature}.
 
The problem of finding an estimate for the matrices $\mm{P}^K$ and $\mm{S}^K$ can therefore be addressed as the problem of finding a set of assignments $\{\vv{u}_{x}^k \Leftrightarrow \pm\vv{u}_{y}^l,1\leq k,l\leq K\}$ based on the comparison of the eigenfunction signatures, namely the histograms. This is an instance of the standard bipartite maximum matching problem whose complexity is $O(K^3)$. Notice however that the eigenvectors are defined up to a sign (hence the presence of matrix $\mm{S}^K$) so  two different histograms $h(\vv{u})$ and $h(-\vv{u})$ can be associated with each eigenfunction and need to be compared. Let $C(h(\vv{u}),h(\vv{v}))$ be a measure of dissimilarity between two histograms.  Computing the dissimilarity of all pairs of eigenfunctions $(\vv{u}_{x}^k,\pm\vv{u}_{y}^l)$ we can build a {\small $K\times K$} matrix $\mm{A}$ whose entries are defined by:
\[
A_{kl} = \min \{ C(h(\vv{u}_x^k),h(\vv{u}_y^l)), C(h(\vv{u}_x^k),h(-\vv{u}_y^l))\}
\]
The Hungarian algorithm provides an optimal solution to the problem of finding an assignment between eigenfunctions of the two graphs (shapes), taking $\mm{A}$ as input and producing as output a permutation matrix. Because the sign ambiguity has been explicitly taken into account, this solves as well for $\mm{S}^K$.
This method provides an good initialization for the point registration method described in section \ref{section:point-registration}.

\begin{figure}[ht]
\centering
\includegraphics[clip=true, viewport= 10 50 500 750, keepaspectratio=true, width=0.88\columnwidth]
{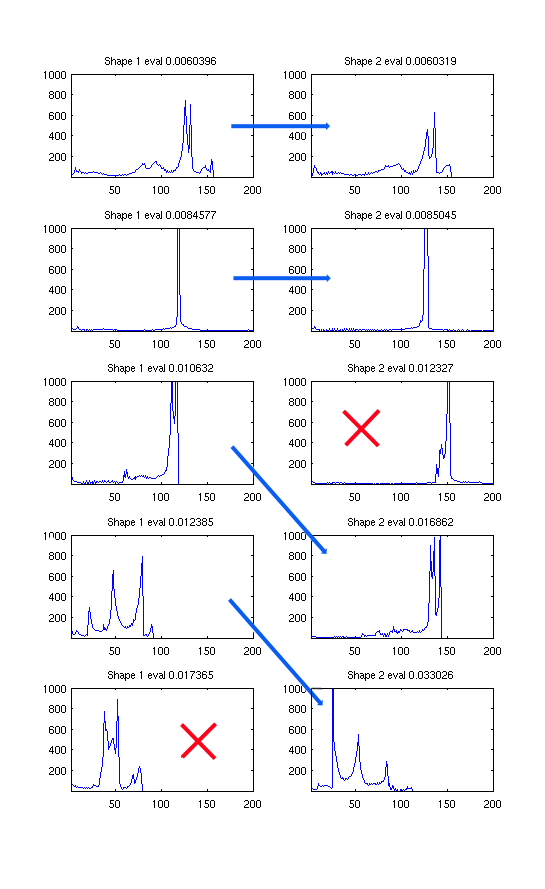}
\vspace{-3mm}\\
\begin{tabular*}{0.45\columnwidth}{@{\extracolsep{\fill}}cc}
(a)& (b)
\end{tabular*}
\caption{Histograms associated with the embeddings of the wooden-mannequin example. \textbf{(a)} pose 1 \textbf{(b)} pose 2. First 5 histograms out of the 20 compared. Notice the sign flip when the fourth histogram on the left is aligned with the fifth histogram on the right (horizontal reflection).}
\label{fig:mann-to-mann-histogram}
\vspace{-5mm}
\end{figure}

As an example, Figure \ref{fig:mann-to-mann-histogram} shows the histograms of the first five eigenfunctions associated with the two different poses of the wooden mannequin of Figure \ref{fig:mann-to-mann-images}. One can visually realize the striking similarity of histograms related to extremely separated poses of the same shape: as a matter of fact the topology of the body is different in the two cases due to the presence of the hands' self-contact. The effectiveness of the histograms as eigenfunction signatures is attested by the correct assignments represented as arrows. Notice that, due to the topological differences, it has not been possible to align all the eigenfunctions. This is not an issue since we only retain the aligned eigenfunctions. These form an eigenspace where finding the assignment is easier. In practice, the algorithm starts with the most significant, e.g. the first 20-25, eigenvalues  and eigenfunctions for each shape, but retains only only the pairs of eigenfunctions with the best alignment score. This improves the robustness of the matching algorithm.

\vspace{-4mm}
\section{Point registration and clustering}
\label{section:point-registration}
\vspace{-3mm}
Points belonging to two poses of the same articulated shape are congruent after graph Laplacian embedding. To find an assignment between them we need to find an optimal alignment between  two congruent sets of points in the reduced space of dimension $K$. An immediate consequence of the result obtained in section~\ref{section:problem-statement} is that articulated shape matching casts into a \emph{point registration} problem, which we propose to solve in the framework of \emph{clustering}. The points $\mathcal{\bar{X}}$ are treated as observations, while the points in the second shape, $\mathcal{\bar{Y}}$, are treated as centers of normally distributed clusters.
In addition to these Gaussian clusters, we will need to consider an {\em outlier component} with uniform distribution. We refer then to our method as {\em unsupervised robust point registration}.

Under these conditions, the likelihood of an observation $x_n$ to belong to a cluster $m$, $1\leq m\leq M$, has a Gaussian distribution with mean \footnote{For simplicity we drop the superscript of $\mm{R}^K$.}  $\mu_{m}=\mm{R}y_m$ and covariance $\mm{\Sigma}$.
We also introduce a $(M+1)^{\text{th}}$ {\em outlier cluster} with uniform distribution. The likelihood of an observation to be an outlier is uniformly distributed over the volume $V$  which contains the embedded shapes. This yields:
\begin{eqnarray*}
&&P(x_n | z_n= m) = \mathcal{N} (x_n | \mm{R}y_m, \Sigma), 1\leq m\leq M,\\
&&P(x_n | z_n = M+1) = \mathcal{U} (x_n | V,0).
\end{eqnarray*}
 One can write the likelihood of an observation as a mixture of these $M+1$ distributions weighted by their priors:
\[
P(x_n) = \sum_{m=1}^{M+1} \pi_m P(x_n | z_n= m).
\]
We introduce the latent variables $\mathcal{Z}=\{z_1,\ldots,z_n\}$, that assign each observation to a cluster. 
By assuming independent and identically distributed observations, we obtain the log-likelihood of the point-set $\mathcal{\bar{X}}$:
\begin{equation}
\ln P(x_1,\ldots,x_N) = \sum_{n=1}^{N} \ln \sum_{m=1}^{M+1} \pi_m P(x_n | z_n= m).
\label{eq:log-likelihood}
\end{equation}
Hence, the problem of point-to-point assignment can be formulated as the maximization of the log-likelihood in eq.~(\ref{eq:log-likelihood}). Due to the presence of the latent variables 
this will be carried out using the EM algorithm. With our notations, EM should solve: 
\begin{equation}
\mm{R}^\ast = \arg \max_{\mm{R}}
E_{\mathcal{Z}}[ \ln P_{\mm{R}}(\mathcal{\bar{X}},\mathcal{Z})| \mathcal{\bar{X}}].
\label{eq:EM}
\end{equation}
That is, maximize the {\em conditional expectation} taken over $\mathcal{Z}$ of the joint log-likelihood (of observations and assignments) {\em given the observations}. The joint likelihood is parameterized by $\mm{R}$. In contrast to standard EM, we estimate a {\em global} orthogonal transformation $\mm{R}$ (instead of $M$ independent $K$-dimensional means), as well as a {\em global} {\small$K\times K$} covariance matrix $\mm{\Sigma}$, constrained to be common to all the clusters (in the interest of avoiding convergence problems when a point $x_n$ is infinitely close to a cluster center $\mm{R}y_m$).

In order to formally derive an analytic expression for the conditional expectation of eq.~(\ref{eq:EM}), we need to compute the posterior class probabilities, namely $P(z_n=m|x_n)=P(z_n=m,x_n)/P(x_n)$. This can be interpreted as the posterior probability of a point $x_n$ to be registered with point $y_m$ and we denote it by $\alpha_{nm}$. We assume that all points $m$ have equal prior probabilities and hence $\pi_1=\ldots=\pi_M=\pi_{in}$. Therefore we have $\pi_{M+1}= 1-M\pi_{in} = \pi_{out}$. Denoting  $d_{\Sigma}$ the Mahalanobis distance\footnote{$d_{\Sigma}(\vv{a},\vv{b}) = (\vv{a-b})\tp \Sigma^{-1}(\vv{a-b})$}, we obtain:
\begin{equation}
\alpha_{nm} = 
\frac{exp(-d_{\Sigma}(x_n,\mm{R}y_m))}
{\sum_{i=1}^{M}  exp(-d_{\Sigma}(x_n,\mm{R}y_i) + \kappa)}.
\label{eq:posteriors}
\end{equation}
The parameter $\kappa$ represents the contribution of the outlier cluster; it is possible to verify that: \hbox{$\kappa=(2\pi)^{K/2}(\det\mm{\Sigma})^{1/2} \pi_{out}/V\pi_{in}$}. Next, we write $P(\mathcal{\bar{X}},\mathcal{Z})$ as:
\[
P(\mathcal{\bar{X}},\mathcal{Z}) = \prod_{n=1}^N \prod_{m=1}^{M+1} \big(
P(x_n|z_n=m)P(z_n=m)
\big)^{\delta_m(z_n)},
\]
where the function $\delta_m(z_n)$ is equal to 1 if $z_n=m$ and to 0 otherwise. When taking the conditional expectation of the log of this likelihood, we need to estimate the conditional expectation of the functions $\delta_m$. One may notice that:
 \[
 E[\delta_m(z_n)|\mathcal{X}] = \sum_{i=1}^{M+1} \delta_m (z_n=i)P(z_n=i|x_n)= \alpha_{nm}.
 \]
 Finally, denoting the current parameter estimate with superscript $(q)$ and dropping out the constant terms, the negative expectation in eq.~(\ref{eq:EM}) can be written as:
 \begin{equation}
 \label{eq:EM-final}
 E(\mm{R}|\mm{R}^{(q)}) =
 \sum_{m=1}^M \xi_m 
 \left(d_{\Sigma} \left( \overline{x}_{m}, \mm{R}y_m\right)
 + \ln \det \Sigma  \vphantom{\sum_{i=1}^N} \right),
 \end{equation}
where, 
\begin{equation}
\begin{array}{ccc}
 \overline{x}_m = \sum_{i=1}^N  \alpha_{im} x_i / \xi_m & \text{and}&
 \xi_m= \sum_i \alpha_{im}.
 \end{array}
 \label{eq:mean-observation}
 \end{equation}
Our algorithm finds one-to-one assignments $\overline{x}_m\Leftrightarrow y_m$. \hbox{$\overline{x}_m$ is a mean} over all $x_i\in\mathcal{\bar{X}}$, built by weighting each $x_i$ with its posterior probability of being assigned to $y_m\in\mathcal{\bar{Y}}$, and normalizing each such assignment by $\xi_m$. In practice, these posteriors ($\alpha_{nm}$) and these weights ($\xi_m$) allow the {\em classification} of the observations into inliers and outliers. For this reason, our algorithm performs {\em unsupervised robust point registration}.

To summarize, the formal derivation outlined above guarantees that the
maximization of eq.~(\ref{eq:log-likelihood}) is equivalent to the
minimization of eq.~(\ref{eq:EM-final})  \cite{McLachlanKrishnan97}. Therefore, one can apply the EM algorithm to eq.~(\ref{eq:EM-final}) as detailed in the next section.

\section{Experimental results}
\label{section:experiments}
\vspace{-3mm}
We applied the method described above to estimate the matching of \hbox{3-D} articulated shapes described as dense sets of voxels. In this section, we show the results of the algorithm on three different objects: a wooden mannequin, a person, and a hand (see Table \ref{table}). We gathered several data sets of poses of these objects using six calibrated cameras and a voxel reconstruction software package. The results show how most of the points are matched in few iterations of the algorithm.

\begin{table}[htb]
{\small
\centering
\begin{tabular}{|c|c|c|c|}
\hline
Example & EM  & Number  & Outliers+ \\
 &  iterations & of voxels & Unmatched\\
\hline
Mannequin & 6 & 12667 - 13848 & 472\\
\hline
Manneq-person  & 8 & 11636 -12267 & 388\\
\hline
Hand & 4 & 13594 -13317 &  39 \\
\hline
\end{tabular}
\caption{Three challenging experiments }}
\label{table}
\vspace{-5mm}
\end{table}

\begin{figure}[ht]
\centering
\small{
\begin{tabular}{cc}
\includegraphics[height=0.42\columnwidth]{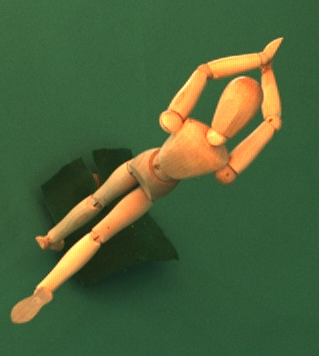} &
\includegraphics[height=0.42\columnwidth]{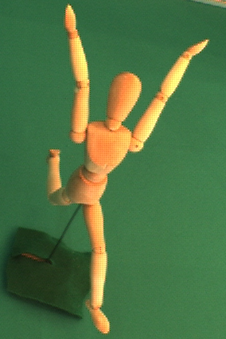}\\
(a) pose 1 & (b) pose 2
\end{tabular}}
\caption{ Two poses of a wooden mannequin. In \textbf{(a)} the hands touch each other introducing important topological differences between the two associated graphs.}
\label{fig:mann-to-mann-images}
\vspace{-3mm}
\end{figure}

\begin{figure}[h]
\centering
\small{
\begin{tabular}{cc}
\includegraphics
[clip=true, viewport= 10  100 760 590, keepaspectratio=true,height=0.28\columnwidth]
{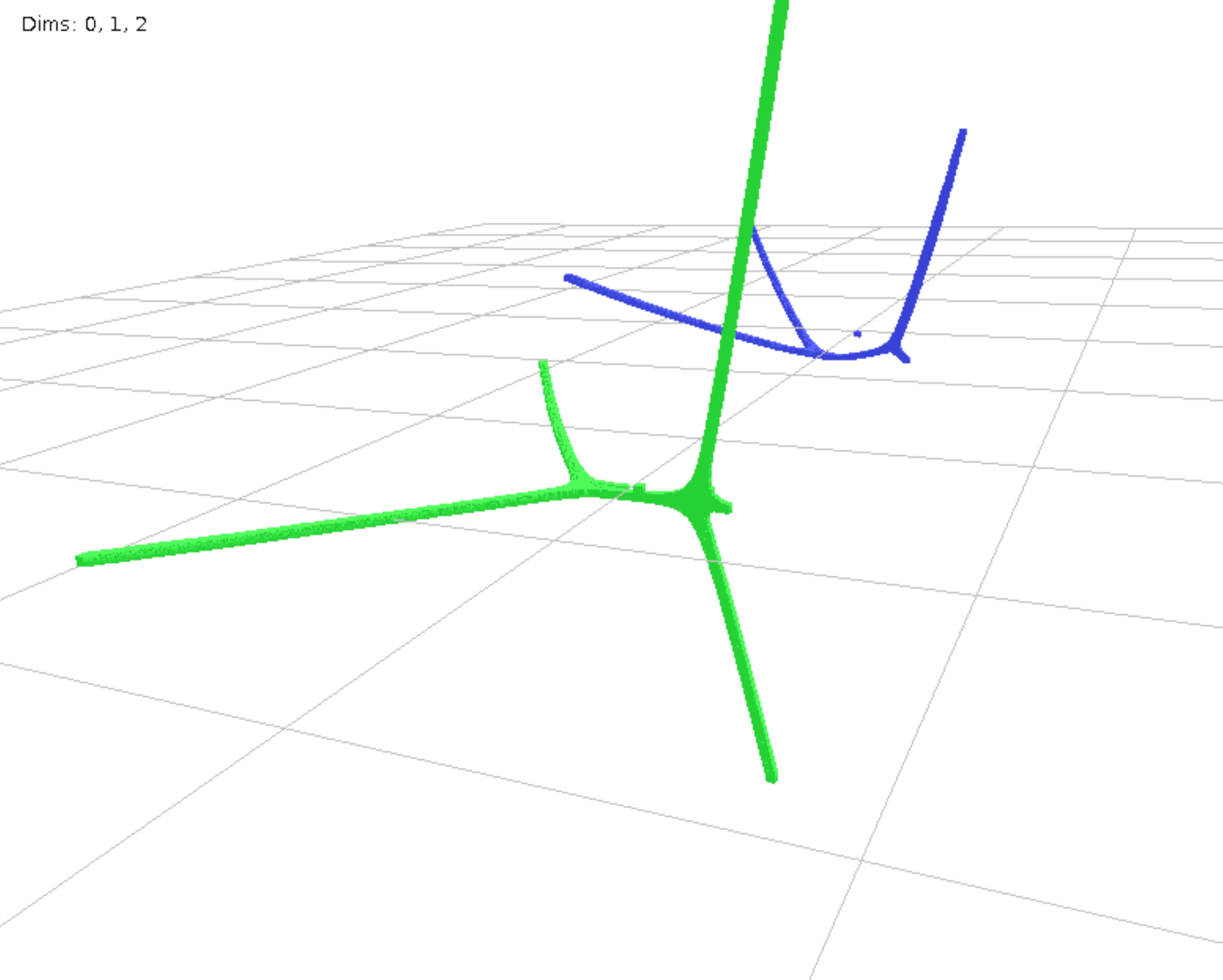} &
\includegraphics
[clip=true, viewport= 0  150 770 590, keepaspectratio=true,height=0.28\columnwidth]
{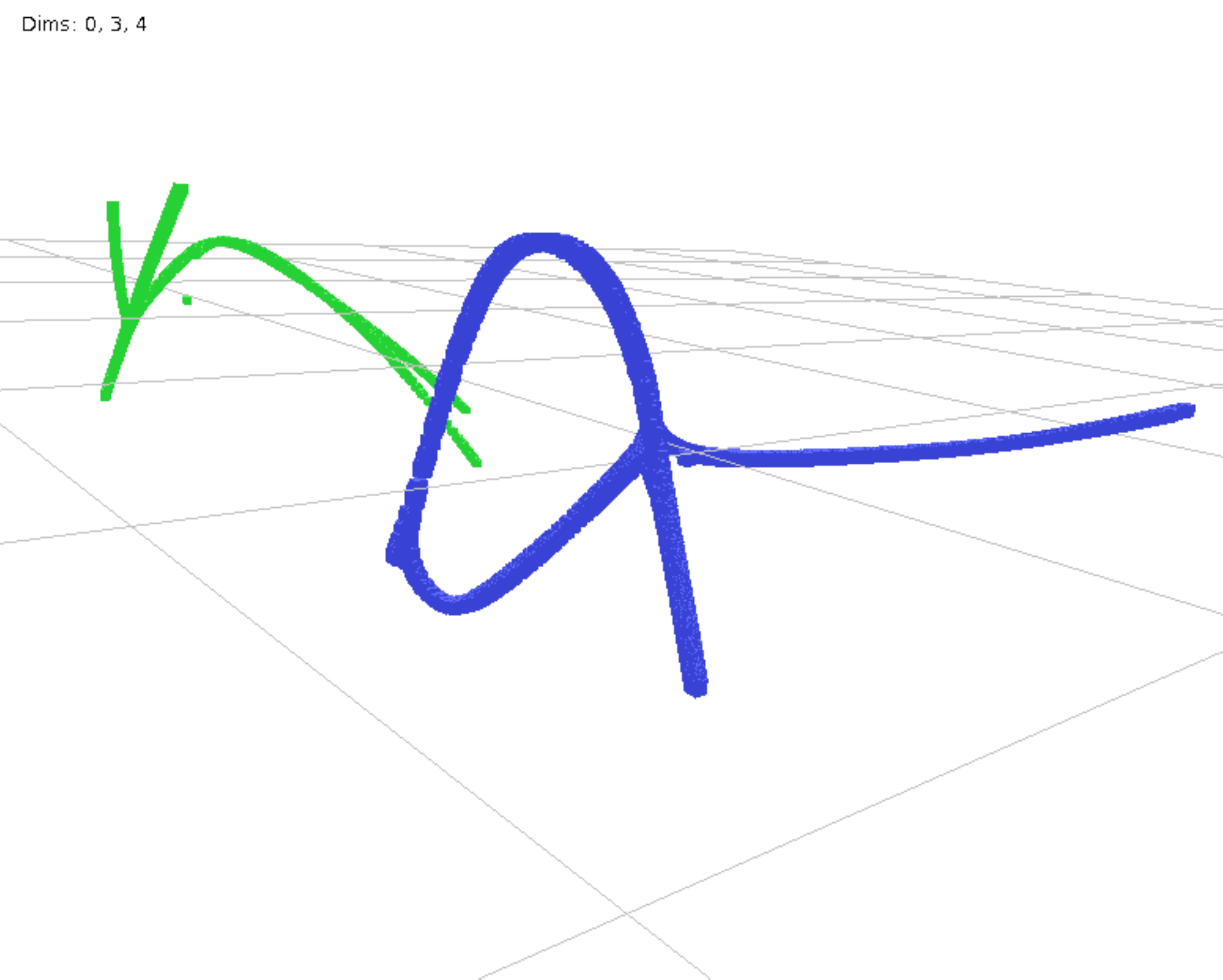}\\
(a) eigenfunctions 1-2-3 & (b) eigenfunction 1-4-5 
\end{tabular}}
\caption{Embeddings before alignment: Embeddings of the two poses of the wooden mannequin in Figure \ref{fig:mann-to-mann-images} (blue: pose 1,  green: pose 2). The embeddings are represented as their three dimensional projections on two different subspaces. \textbf{(a)} subspace spanned by eigenfunctions1-2-3, \textbf{(b)} subspace 1-4-5.}
\label{fig:mann-to-mann-embed-init}
\vspace{-2mm}
\end{figure}

\begin{figure}[!h]
\centering
\includegraphics[width=0.77\columnwidth]
{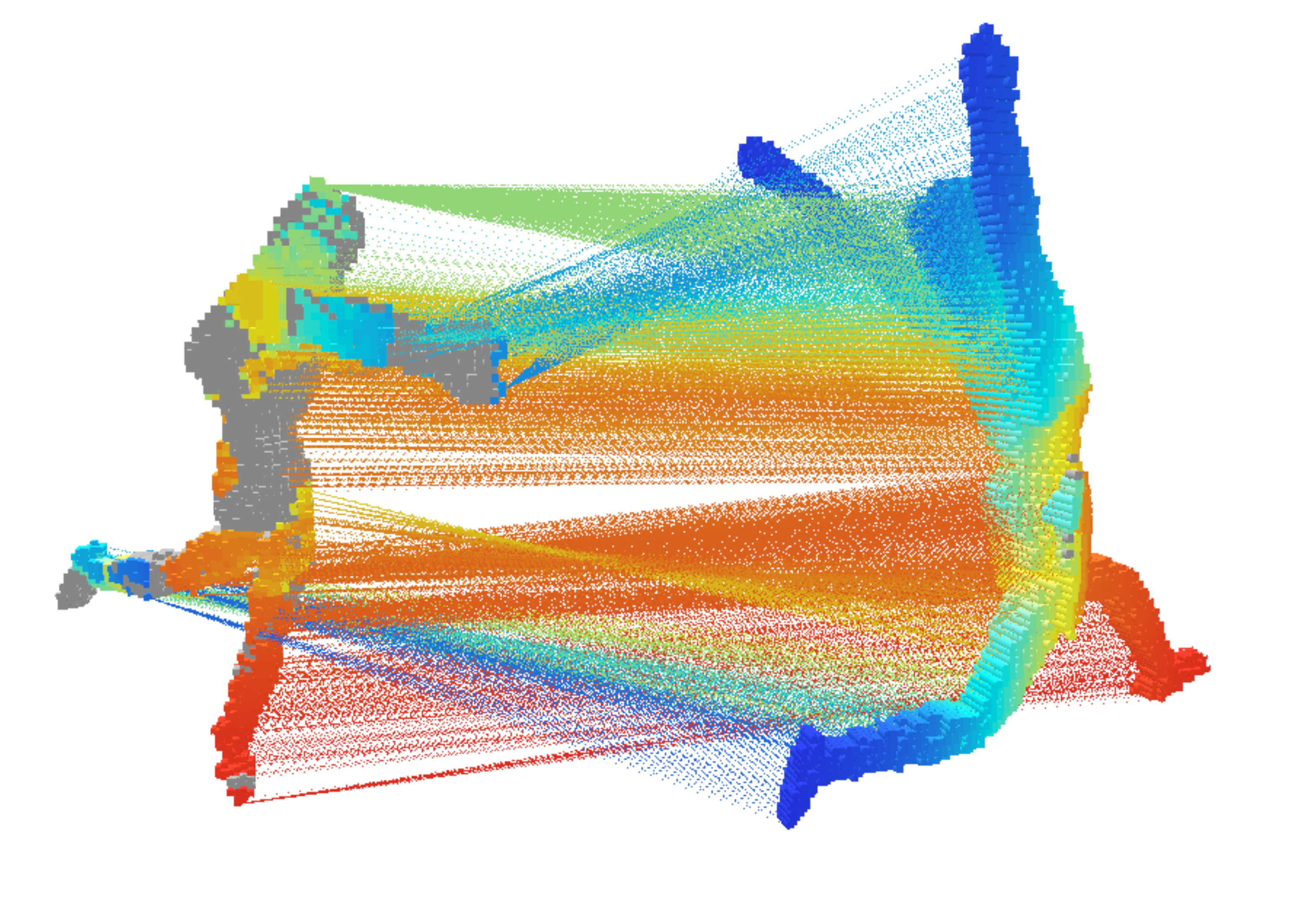}
\\ \vspace{-4mm}{\small (a) After eigenfunction alignment (EM initialization)} \\
\includegraphics[width=0.77\columnwidth]
{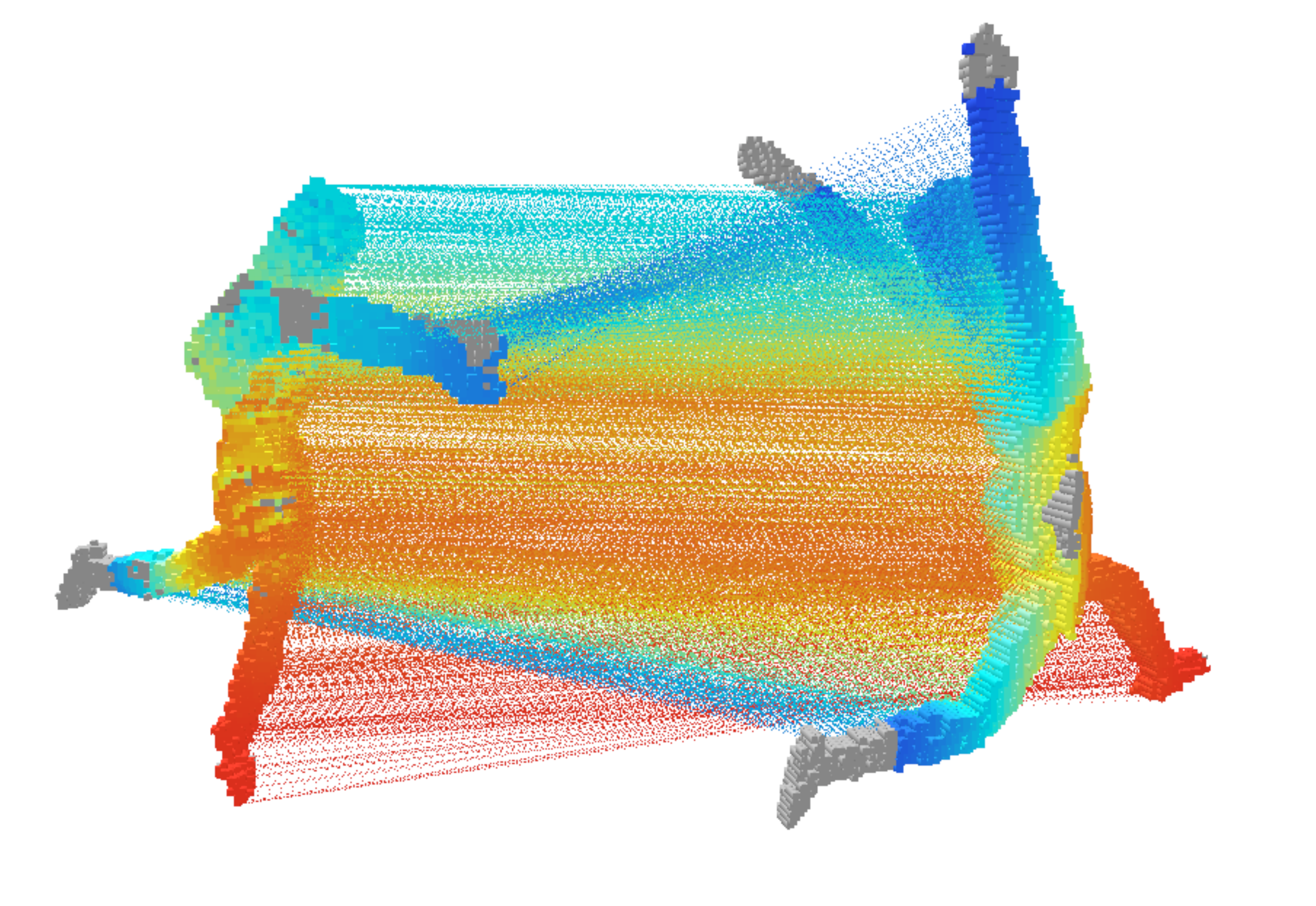}
\\ \vspace{-4mm}{\small (b) After EM convergence} 
\caption{Dense match between the two poses  of the wooden mannequin in Fig. \ref{fig:mann-to-mann-images}. \textbf{(a)} Assignment found after aligning the eigenspaces with the eigenfunction signature histograms. \textbf{(b)} Final result after convergence of EM. Points in grey illustrate outliers.}
\label{fig:mann-to-mann-EM}
\vspace{-6mm}
\end{figure}

Figure \ref{fig:mann-to-mann-images} shows two widely separated poses of the same articulated object, a wooden mannequin. The algorithm starts by building  the Laplacian matrices of these shapes in order to map them onto a lower dimensional space (Figure \ref{fig:mann-to-mann-embed-init}). Then, the signatures of the associated eigenfunctions are computed and matched as described in Section \ref{section:laplacian-eigenfunctions} in order to find a permutation matrix between the eigenvectors associated with the two embeddings and solve for the sign ambiguity (see Figures \ref{fig:mann-to-mann-histogram} and \ref{fig:mann-to-mann-EM}-a).
In this experiment, the dimension of the embedding space turned out to be $K=5$. 
Notice how the contact between the two hands alters the topology of the embedded shape (Figure \ref{fig:mann-to-mann-embed-init}); without the eigenfunction matching stage, i.e. using only eigenvalue to order the eigenfunctions, the point matching becomes a difficult task and the EM algorithm would easily fall into a local minimum.

\begin{figure}[!h]
\centering
\small{
\begin{tabular}{cc}
\includegraphics
[clip=true, viewport= 10  100 636 520, keepaspectratio=true,height=0.28\columnwidth]
{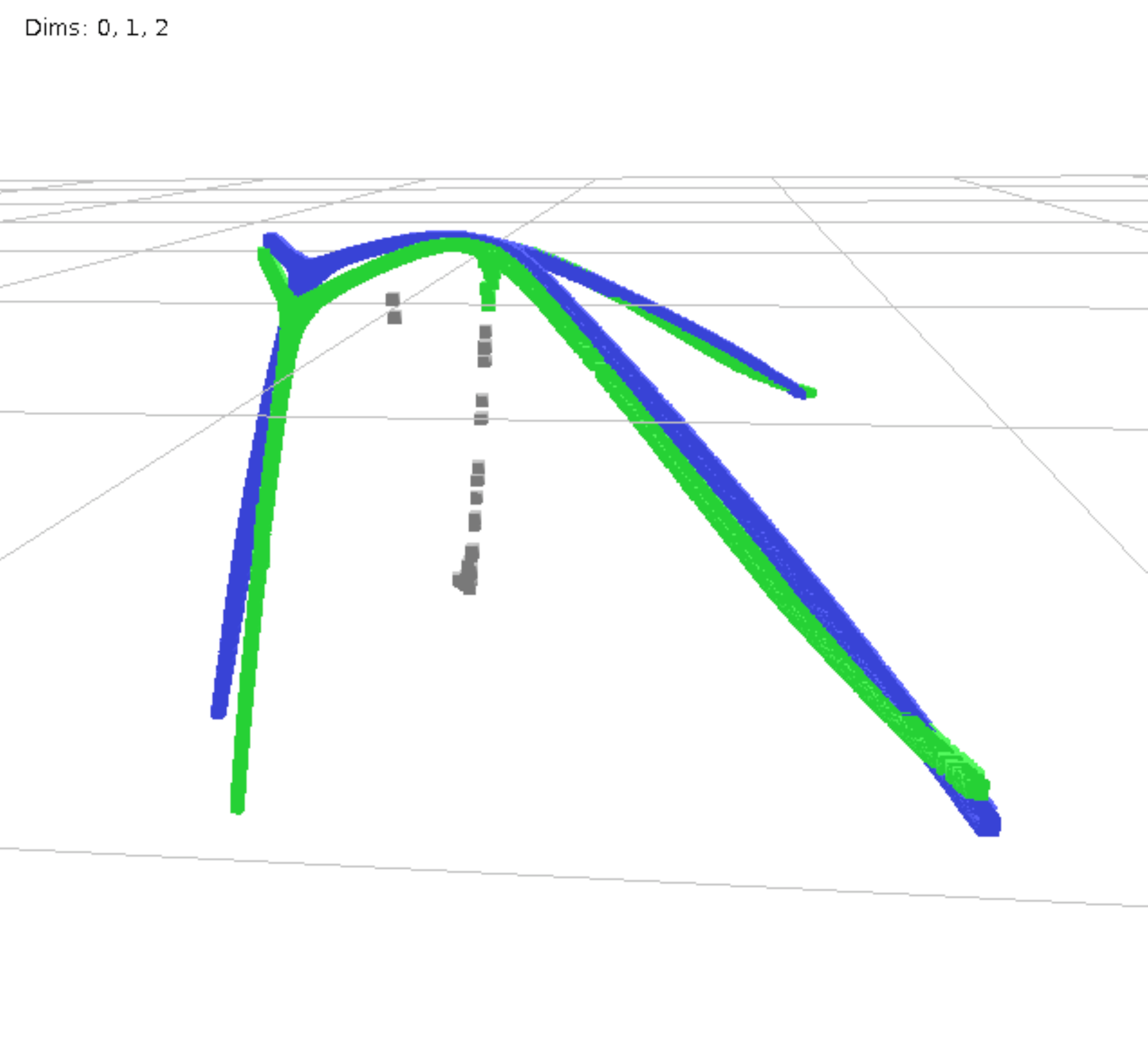} &
\includegraphics
[clip=true, viewport= 10  100 636 553, keepaspectratio=true,height=0.28\columnwidth]
{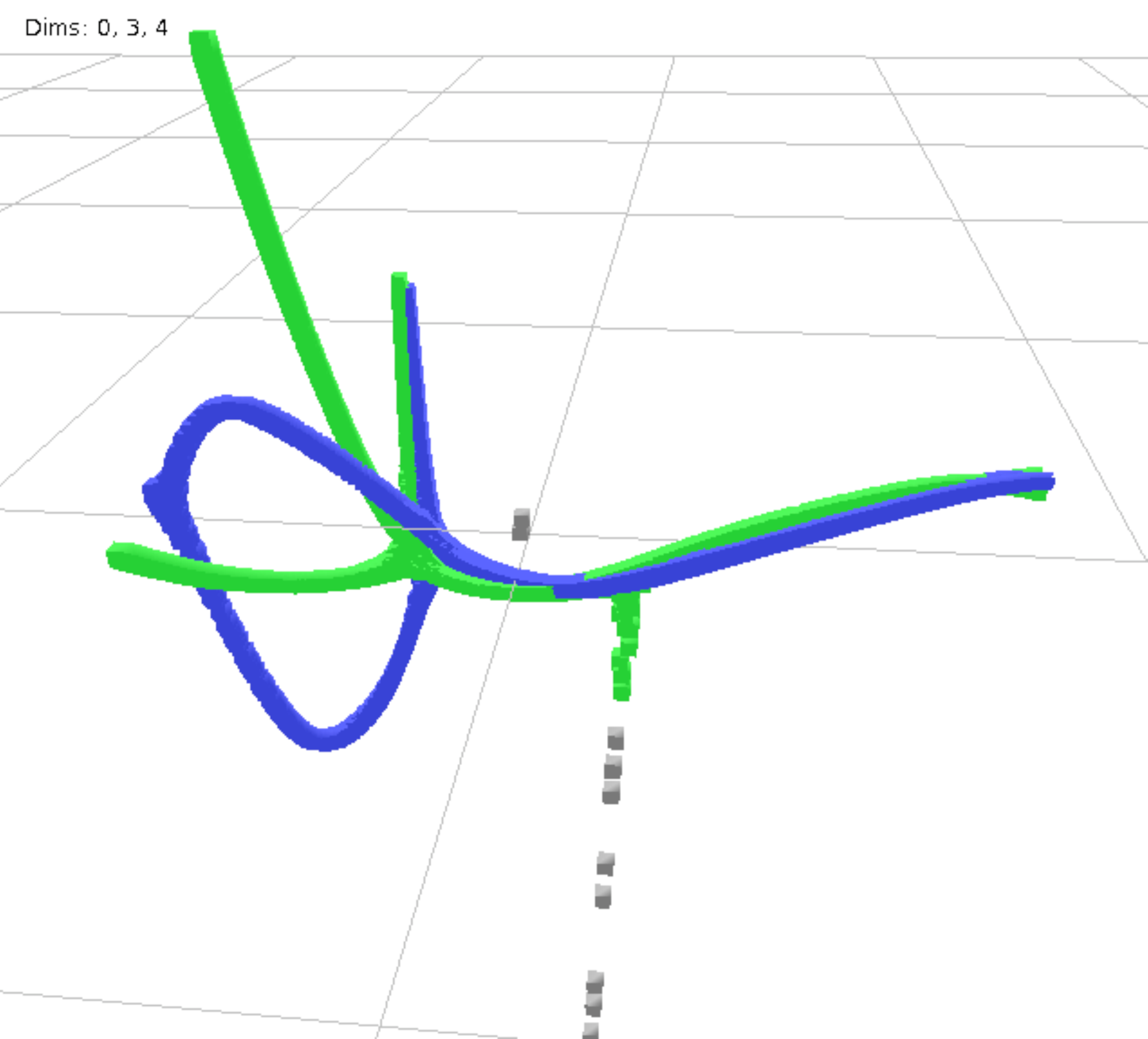}\\
(a) eigenfunctions 1-2-3 & (b) eigenfunction 1-4-5 
\end{tabular}}
\caption{Aligned embeddings: result after eigenfunction alignment and EM. Only the $K$ retained eigenfunctions are shown.}
\label{fig:mann-to-mann-embed-aligned}
\vspace{-4mm}
\end{figure}

Then, the algorithm performs point registration under a {\small $K\times K$} orthogonal transformation, using the EM algorithm described in Section \ref{section:point-registration}. This procedure treats the first point-set as observations and the second point-set as cluster centers. EM needs to be initialized, by provision of an initial orthogonal transformation $\mm{R}$ and a covariance matrix $\mm{\Sigma}$, as well as values for the cluster priors $\pi_1\ldots\pi_M$ (see below). We use the eigenfunction alignment scheme of Section \ref{section:laplacian-eigenfunctions} to provide an initialization for $\mm{R}$. We also use an isotropic covariance $\sigma^2\mm{I}$ for the initialization, where $\sigma$ is chosen sufficiently large to allow evolution of the algorithm and $\mm{I}$ is the $K\times K$ identity matrix. In practice the initial covariance is of the order of 15 voxels. The cluster priors are chosen such that $\pi_1=\ldots=\pi_M=\pi_{in}$ and $\pi_{M+1}= 1-M\pi_{in} = \pi_{out}$. Moreover $\pi_{in}$ is chosen as a small proportion of the working volume $V$.

The \textit{E-step} computes the posterior probabilities associated with each observation using eq. (\ref{eq:posteriors}); in addition to the current estimates of $\mm{R}$ and $\mm{\Sigma}$, one needs to specify the uniform-component parameter $\kappa$. With the explained choice for $\pi_{in}$ and $\pi_{out}$  we have $\kappa \propto (\det \mm{\Sigma})^{1/2}$. The \hbox{\textit{M-step}} estimates the transformation $\mm{R}$ and the covariance $\mm{\Sigma}$. The estimation of $\mm{R}$ is done in closed form using an extension of \cite{ArunHuangBlonstein91} to deal with orthogonal matrices of arbitrary dimension rather than with \hbox{3-D} rotations. The covariance matrix can be easily estimated from eq. (\ref{eq:EM-final}) as is classically done \cite{Bishop2006}. At convergence, EM provides an optimal value for the maximum likelihood, as well as final estimates for the posterior probabilities. This posteriors assign each ``mean'' observation $ \overline{x}_m$ (eq. \ref{eq:mean-observation}) to a ``cluster center'' $y_m$.

Figure \ref{fig:mann-to-mann-embed-aligned} shows the final alignment of the two embedded shapes of Figure \ref{fig:mann-to-mann-embed-init}. Even though the hands self-contact induces significant differences in the graph topology and thus in the embedded shapes, there is still a significant set of voxels which preserves the same structure in both poses. Our sub-graph matching algorithm is capable to recover this maximal subset and successfully match the point-sets, despite the presence of a large number of unmatchable points. This performance is due the Laplacian embedding, but mainly to the outlier rejection mechanism introduced in Section \ref{section:point-registration}, which rejects this unmatched points to a class of their own and prevents them to influence the correctness of the orthogonal alignment.

Figure \ref{fig:mann-to-mann-EM} shows the registered sets of voxels after the eigenfunction alignment and EM. It is interesting to see how the initial registration, based on aligning the eigenfunctions of the Laplacian, is already capable to provide a good assignment for the voxels in the limbs. This has a simple justification in terms of spectral clustering, since each eigenfunction corresponds to a well-identified group of voxels.
\vspace{-4mm}
\begin{figure}[hb]
\centering
{\small 
\begin{tabular}{cc}
\includegraphics[height=0.45\columnwidth]{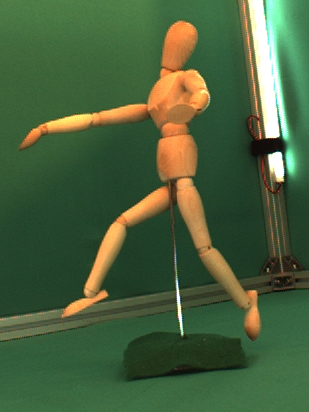} &
\includegraphics[height=0.45\columnwidth]{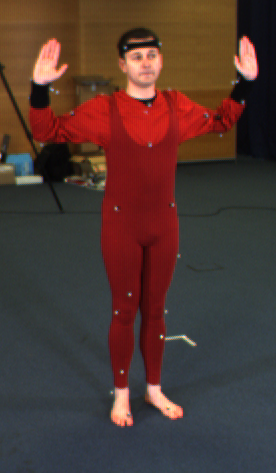}\\
(a) Shape 1 & (b) Shape 2
\end{tabular}
\includegraphics
[clip=true, viewport= 10 30 725 510, keepaspectratio=true,width=0.8\columnwidth]
{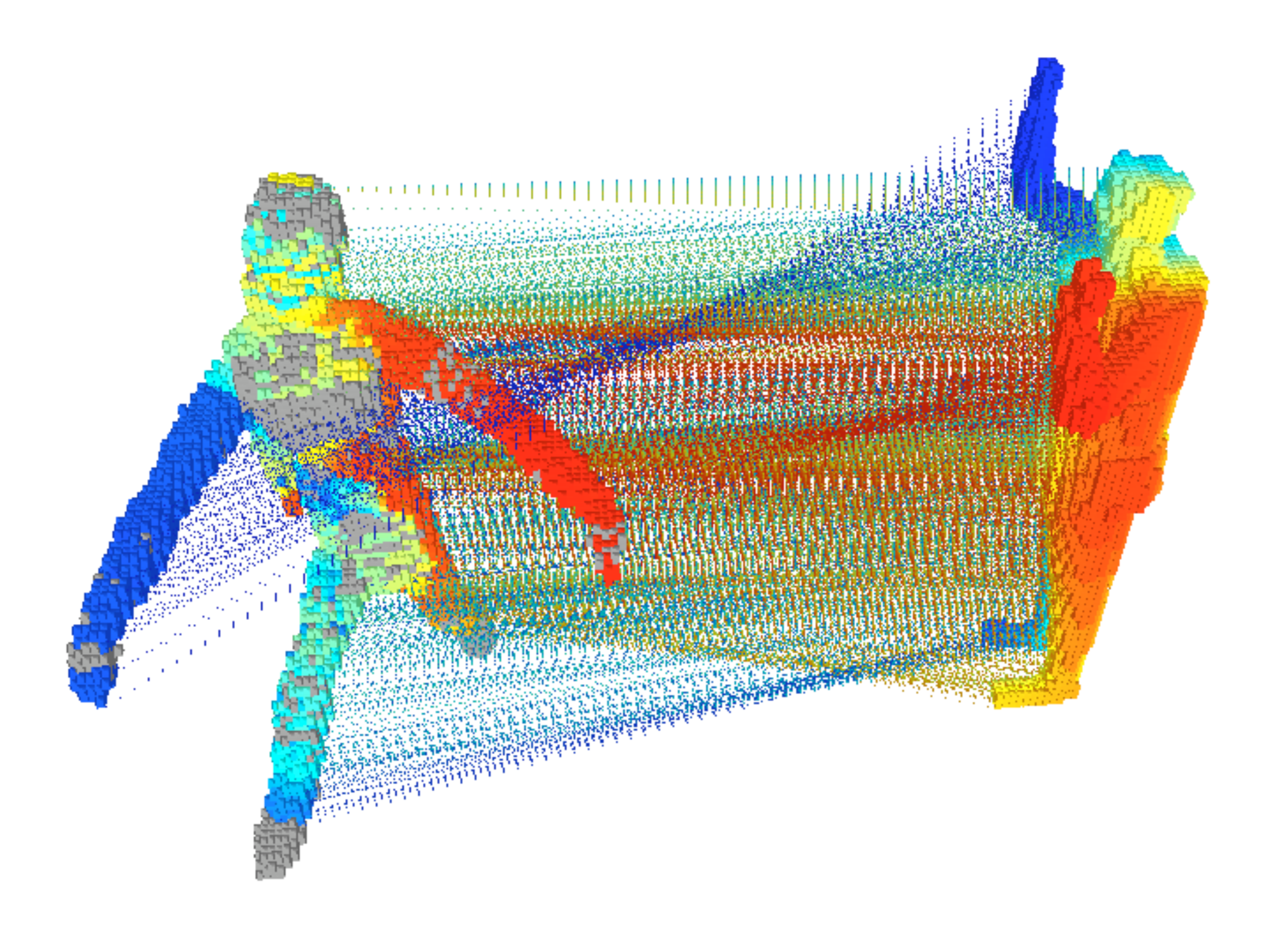}
\\ (c)  Match between (a) and (b)\\
}
\caption{Matching the wooden mannequin to a person is possible in our framework, as the maximal sub-graph isomorphism is sought. As long as two shapes have a common sub-structure, the algorithm is normally able to find it.}
\label{fig:mann-to-human-images}
\vspace{-5mm}
\end{figure}

Figure \ref{fig:mann-to-human-images} illustrates the capability of the proposed algorithm to handle  shapes of \emph{different datasets}. It shows the matching between the wooden mannequin and a person. Even though they have roughly the same ``shape'' (described as a graph whose nodes are voxels), the relative dimensions of the limbs are not the same, the shape of the rigid links forming them is also altered, and finally they are sampled with a different number of voxels. Nevertheless our methodology delivers remarkable results.

\begin{figure}[ht]
\centering
{\small 
\begin{tabular}{cc}
\includegraphics[height=0.4\columnwidth]{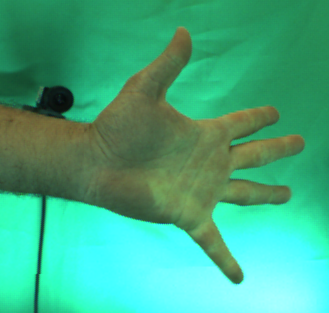} &
\includegraphics[height=0.4\columnwidth]{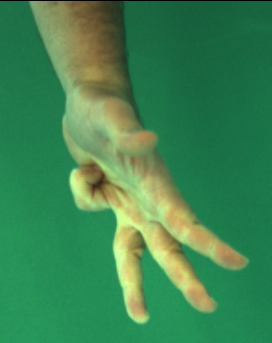}\\
(a) Pose 1 & (b) Pose 2 
\end{tabular}
\includegraphics[width=0.75\columnwidth]{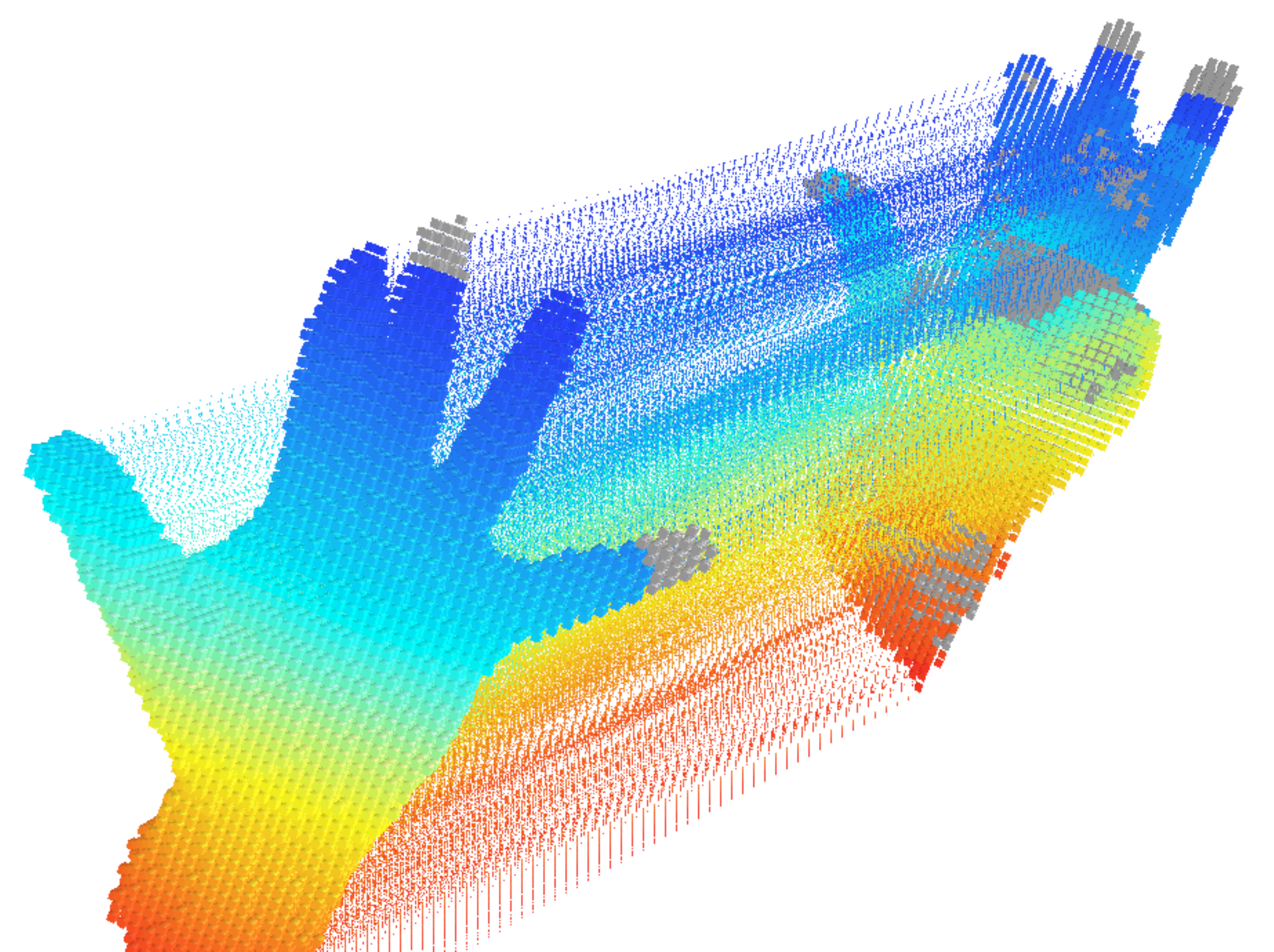}\\
(c) Match between (a) and (b)}
\caption{Matching two different poses of a hand showing bending and a different type of self-contact. Points in the extended finger are matched to points in the bent finger. The unmatched region in the palm reflects the local change of the connectivity.}
\label{fig:hand-to-hand}
\vspace{-4mm}
\end{figure}

In a final  test, Figure \ref{fig:hand-to-hand} shows an alignment between two different poses of a hand. The bending of the small finger creates a different type of self-contact. The algorithm solves for most of the matches and in particular finds appropriate matchings for the small finger. The unmatched region in the palm reflects the change in the connectivity between the palm and the finger which alters the embedded shapes.

\subsection{Discussion}
\vspace{-3mm}
The good performance of the algorithm is the result of several choices. First, the use of voxel-sets as input for our experiments instead of other shape representations such as meshes; indeed, due to the statistical properties of our large datasets, the voxel representations leads to highly stable embeddings,  they are easier to generate and to maintain than surfaces, and allow to exploit the full volumetric information of the shape. Furthermore, the fact that they are regularly sampled contributes to the convergence of the Laplacian embedding towards a ``geometric-aware'' eigenbasis \cite{BelkinNiyogi2003}. Second, the definition of a \textit{local} graph-connectivity allows the treatment of the otherwise difficult cases related to self-contacts and topology changes. These cannot be handled when completely connected graphs are defined since a self-contact would imply changes in all the pairwise distances. Furthermore, locality gives rise to very sparse Laplacian matrices and thus to an efficient calculation of the eigendecomposition (we use {\small ARPACK}). Additionally, the initialization is usually good for EM to converge in a few iterations. As a consequence,  the procedure is time-efficient, and that whole matching procedure of large voxel-sets takes only a few seconds.
Finally, it is important to mention that our method can be applied to other type of data since it relies only on geometric cues and does not use the photometric information as suggested in \cite{hilton:iccv2007}.
\vspace{-2mm}
\section{Conclusions}
\label{section:conclusions}
\vspace{-3mm}
This paper describes a new method for establishing dense correspondences between the points associated with two articulated objects. We address the problem using both spectral matching and unsupervised point registration. We formally introduce graph isomorphism using the Laplacian matrix, and we provide an analysis of the matching problem whenever the number of nodes in the graph (or, equivalently the number of voxels in the shape) is very large, i.e. of the order of $10^4$. We show that there is a simple equivalence between graph isomorphism and point registration under the group of orthogonal transformations, when the dimension of the embedding space is much smaller than the cardinality of the point-sets.

The eigenvalues of a large sparse Laplacian cannot be reliably ordered. We propose an elegant alternative to eigenvalue ordering, using eigenfunction histograms and alignment based on comparing these histograms. The point registration that results from eigenfunction alignment yields an excellent initialization for the EM algorithm,  subsequently used only to refine the registration. Indeed, each eigenfunction corresponds to a ``mode'' within the framework of spectral clustering.

Moreover, the proposed EM algorithm differs significantly from previously proposed methods: it leads to a general purpose unsupervised robust point registration method. The algorithm can deal with large discrepancies between the two sets of points, by incorporating a uniform component in the mixture model. Eventually, it assigns an optimal ``mean'' observation to each cluster center, which amounts to a dense point-to-point matching.

In the future we plan to study more thoroughly the link between spectral matching and spectral clustering. We believe that this may solve the problem of dimension selection for the embedding space and that this is equivalent to choosing the number of modes in spectral clustering algorithms. The combination of spectral matching with probabilistic methods, along the lines described in this paper, can be applied to many other representations, such as \hbox{3-D} meshes, \hbox{2-D} silhouettes,  bags of features, and so forth.
\vspace{-3mm}
{\small
\bibliographystyle{ieee}

}

\end{document}